\documentclass[runningheads]{llncs}
\usepackage{amsfonts,amssymb,amsmath}
\usepackage{threeparttable}
\usepackage{multirow}
\usepackage{booktabs}
\usepackage{marvosym}
\usepackage{color}
\usepackage{graphicx}
\usepackage{array}
\usepackage{algorithm}               
\usepackage{algorithmic}             
\usepackage{multirow}                
\usepackage{bm}
\usepackage{footmisc}
\usepackage{makecell}
\usepackage[table,xcdraw]{xcolor}
\usepackage{lipsum}
\newcommand\blfootnote[1]{%
\begingroup
\renewcommand\thefootnote{}\footnote{#1}%
\addtocounter{footnote}{-1}%
\endgroup
}
\begin{document}
\title{A Benchmark for Weakly Semi-Supervised Abnormality Localization in Chest X-Rays}
\titlerunning{Point Beyond Class}
%
\author{Haoqin Ji$^{\dagger 1,2,3}$, Haozhe Liu$^{\dagger 1,2,3}$, Yuexiang Li$^{\dagger 3}$, Jinheng Xie$^{1,2}$, Nanjun He$^{\textrm{\Letter} 3}$, Yawen Huang$^{3}$, Dong Wei$^{3}$, Xinrong Chen$^4$, Linlin Shen$^{\textrm{\Letter} 1,2}$, Yefeng Zheng$^{3}$}
\institute{$^1$Computer Vision Institute, College of Computer Science and Software Engineering,\\
 $^2$ AI Research Center for Medical Image Analysis \& Diagnosis, \\
Shenzhen University, \\\email{llshen@szu.edu.cn} \\
$^3$ Jarvis Lab, Tencent, \\ \email{nanjunhe91@163.com} \\
$^4$ Academy for Engineering and Technology, Fudan University}
%
%
%
\maketitle              
\blfootnote{$\dagger$ Equal Contribution}
\begin{abstract}
    Accurate abnormality localization in chest X-rays (CXR) can benefit the clinical diagnosis of various thoracic diseases. However, the lesion-level annotation can only be performed by experienced radiologists, and it is tedious and time-consuming, thus difficult to acquire. Such a situation results in a difficulty to develop a fully-supervised abnormality localization system for CXR. In this regard, we propose to train the CXR abnormality localization framework via a weakly semi-supervised strategy, termed Point Beyond Class (PBC), which utilizes a small number of fully annotated CXRs with lesion-level bounding boxes and extensive weakly annotated samples by points. Such a point annotation setting can provide weakly instance-level information for abnormality localization with a marginal annotation cost. Particularly, the core idea behind our PBC is to learn a robust and accurate mapping from the point annotations to the bounding boxes against the variance of annotated points. To achieve that, a regularization term, namely multi-point consistency, is proposed, which drives the model to generate the consistent bounding box from different point annotations inside the same abnormality. Furthermore, a self-supervision, termed symmetric consistency, is also proposed to deeply exploit the useful information from the weakly annotated data for abnormality localization. Experimental results on RSNA and VinDr-CXR datasets justify the effectiveness of the proposed method. When $\leq$20\% box-level labels are used for training, an improvement of $\sim$5\% in mAP can be achieved by our PBC, compared to the current state-of-the-art method (\emph{i.e.,} Point DETR). Code is available at \url{https://github.com/HaozheLiu-ST/Point-Beyond-Class}.
\end{abstract}
%
%
%
\section{Introduction}
As a noninvasive diagnostic imaging examination, chest X-rays are widely used for screening various thoracic diseases \cite{luo2021oxnet,wang2017chestx}. Radiologists routinely need to screen hundreds CXRs per day, which are extremely laborious. To alleviate the workload of radiologists, an accurate automated abnormality localization system for CXRs is worthwhile to develop.
With the recent advances in deep neural networks \cite{Liu_2021_ICCV,goodfellow2016deep,vaswani2017attention,he2016deep}, numerous modern object detectors \cite{law2018cornernet,lin2017feature,lin2017focal}, such as FCOS \cite{tian2019fcos} and Faster R-CNN \cite{ren2015faster,ren2016faster}, have been proposed, which can be adopted for abnormality localization. However, training these detectors often requires extensive data annotated with lesion-level bounding boxes. 
Such lesion-level annotations are difficult to acquire, since the annotation process yields an over-heavy workload for radiologists. 
Therefore, reducing the annotation cost gradually becomes the core challenge for the development of automated CXR abnormality localization frameworks.

\vspace{-0.1cm}
\begin{figure}[!tb]
    \centering
    \includegraphics[width=0.85\textwidth]{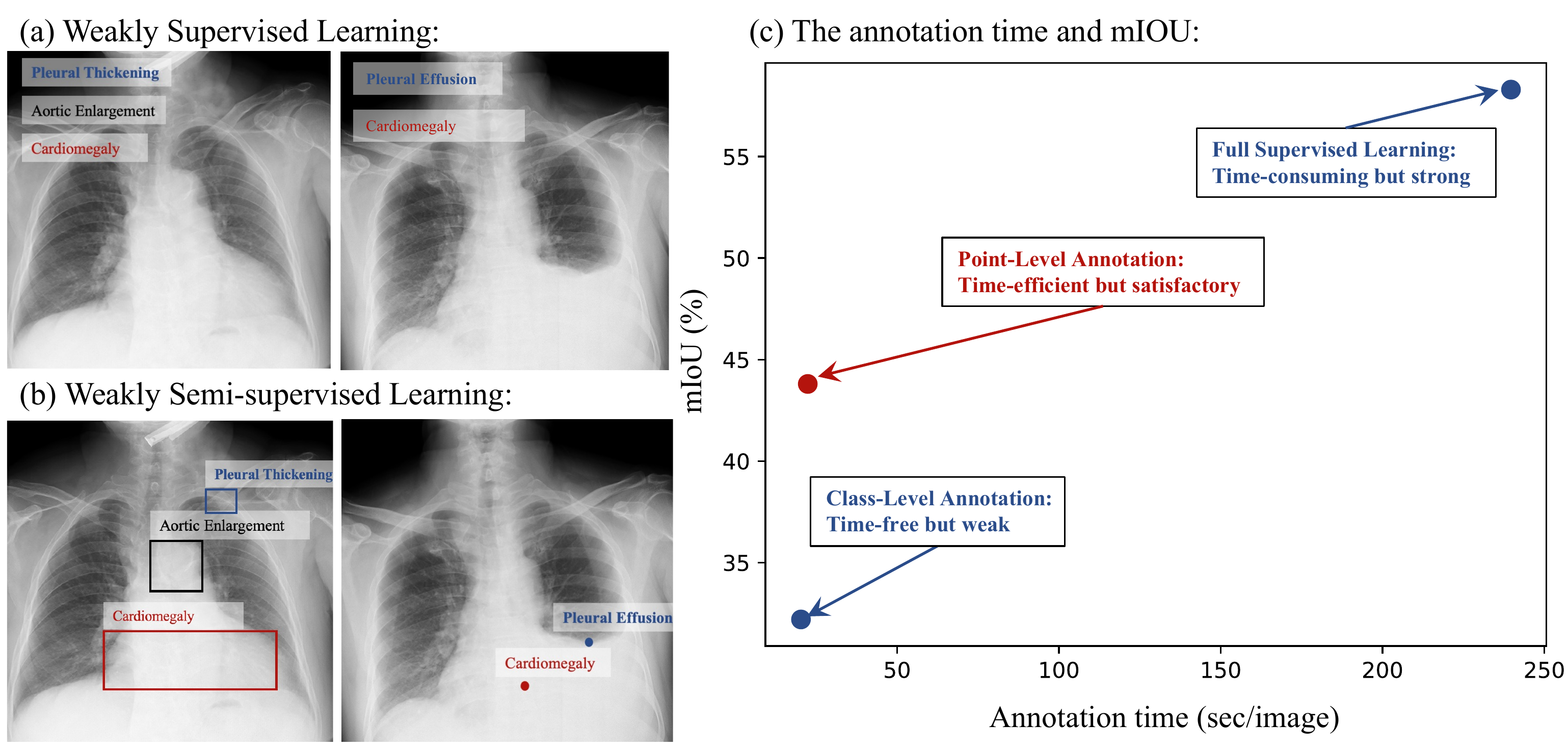}
    \vspace{-0.3cm}
    \caption{Our solution to reduce the cost of annotation for abnormality localization in chest X-Rays. (a) is a general solution, i.e., weakly supervised learning, which only utilizes class-wise annotation for lesion detection. Compared to weakly supervised learning (a), the proposed solution (b) adopts limited bounding boxes and extensive point-level labels to train the detector. Based on the analysis of annotation time and performance (c) carried on PASCAL VOC \cite{bearman2016s}, point-level annotation do not significantly increase the time cost but improve performance effectively \cite{bearman2016s}. }
    \label{fig:motivation}
    \vspace{-2mm}
\end{figure}

To address such a problem, 
various weakly supervised learning methods \cite{wang2021cpnet,bilen2015weakly,Xie_2022_CVPR,Xie_2022_CVPR_1,Xie_2021_ICCV} have been proposed.
Concretely, the weakly supervised object detection methods adopt the data with weak annotations, instead of lesion-level bounding boxes, for network training.
As shown in Fig. \ref{fig:motivation} (a), a typical solution for weakly supervised object detection is using the image-level annotations. These image-level-annotation-based methods localize the objects through region proposals \cite{bilen2015weakly}, which generally depend on the boundary of the objects. However, the boundaries of lesions in chest X-ray images are commonly not clear; therefore, massive invalid proposals may be generated by these image-level-annotation-based methods. Due to this reason, current
weakly supervised object detection methods could not achieve competitive  performance against the fully-supervised counterparts. 
As shown in Fig. \ref{fig:motivation} (c), the empirical study carried by Bearman \emph{et al.} \cite{bearman2016s} is a solid evidence. 
The model can only achieve a mean IoU of $32.2\%$ under the weakly-supervised setting with the image-level annotations, which is boosted to $58.3\%$ by switching to the fully-supervised strategy.
Recent study \cite{Chen_2021_CVPR} proposed a new setting, called weakly semi-supervised object detection (WSSOD), which may be a potential solution for the problem of invalid proposals occurring in current weakly supervised approaches based on image-level annotation.
Concretely, as shown in Fig. \ref{fig:motivation} (b), a novel annotation (\emph{i.e.,} a point inside the object) was adopted to train the object detector together with a small number of fully-labeled samples. According to the reported result \cite{bearman2016s} shown in Fig. \ref{fig:motivation} (c), such a weakly semi-supervised setting can significantly improve the performance of object detection with a marginal extra annotation cost,
since the point-level annotation can provide weakly instance-level information.


In this paper, we evaluate the effectiveness of weakly semi-supervised learning strategy for abnormality localization task with chest X-rays, and accordingly establish a publicly-available benchmark by including various baselines, \emph{e.g.,} image-level-annotation-only weakly supervised, semi-supervised and fully-supervised approaches.
While transferring the existing WSSOD framework (Point DETR \cite{Chen_2021_CVPR}) to CXR abnormality localization task, we notice that the framework is very sensitive to the positions of point annotations. This is because the semantic information contained in the center and boundary of lesion area is different---the points closer to the center provide more useful information for the generation of pseudo bounding boxes. To this end, we propose a regularization term, namely multi-point consistency, to enforce the detector to yield a consistent pseudo bounding box for different points locating in the same lesion area. Furthermore, a self-supervised constraint, termed symmetric consistency, is also proposed to more reasonably explore the weakly annotated data and accordingly boost the abnormality localization performance. Integrating the proposed multi-point consistency and symmetric consistency into the Point DETR \cite{Chen_2021_CVPR}, we form a 
new framework, namely Point Beyond Class (PBC), for abnormality localization with CXRs. Experimental results on publicly available CXR datasets show that the proposed PBC method can significantly outperform all the baselines. 


\section{Revisit of Point DETR}
In this section, we firstly review the pipeline and the network architecture of Point DETR \cite{Chen_2021_CVPR}, and then analyze the challenges unsolved in this scheme. 

\vspace{1.5mm}
\noindent {\bf Pipeline.} Referring to the Point DETR, the conventional WSSOD for abnormality localization in chest X-rays can be represented as a multi-stage scheme: 1) Train a teacher model with a small number of CXRs labeled with both points (randomly selected inside the boxes) and lesion-level bounding boxes; 2) Generate pseudo bounding boxes for the CXRs with point-level annotations only using the well-trained teacher model; and 3) Train a student detector by utilizing the samples with ground truth box labels and the ones with pseudo labels.

\vspace{1.5mm}
\noindent{\bf Network Structure.} The Point DETR adopts a point encoder $\mathcal{F}_p(\cdot)$ to embed the point annotation $\mathbb{X}_p$ into a latent space for object query. Specifically, $\mathcal{F}_p(\cdot)$ decomposes $\mathbb{X}_p$ into position $(x,y) \in [0, 1]^2$ and class-wise annotation $c$. By utilizing a fixed positional encoding method \cite{vaswani2017attention,parmar2018image,carion2020end}, $(x,y)$ can be transferred to a $q$-dimensional code vector $\mathbf{V}_p \in \mathbb{R}^{q}$, while we predefine a learnable embedding code vector with the same dimension ($\mathbf{V}_c \in \mathbb{R}^{q}$) for the category information $c$. Hence, the object query $\mathbf{V}_f$ can be formulated as: $\mathbf{V}_f = \mathbf{V}_p + \mathbf{V}_c.$
Apart from the point encoder, the Point DETR has an image encoder, consisting of a convolutional neural network (CNN) and a Transformer encoder, to encode the CXR images. Concretely, the image encoder firstly embeds the image to feature maps via the CNN. Then, the feature maps are flattened with the positional encoding and fed to the Transformer encoder. The object query $\mathbf{V}_f$ is attended to the image features extracted by image encoder via a Transformer decoder. After that, the output of Transformer decoder is sent to a shared feed forward network (FFN) \cite{carion2020end} for bounding box prediction.

Although Point DETR achieved a satisfactory performance on object detection in natural images, there are still some challenges unsolved for adapting the framework for lesion localization in chest X-rays: 

\vspace{1.5mm}
\noindent In {\bf step 1}, since {there is less contextual information contained in gray-scale CXRs, compared to the natural color images}, the performance of teacher model might be affected by the positions of point-annotations. Specifically, the points closer to the center of lesion area provides more useful information for pseudo label generation than the ones locating around the lesion boundary.

\vspace{1.5mm}
\noindent In {\bf step 2}, the images with point-level annotations are only employed to generate pseudo labels in the pipeline of Point DETR. The rich information contained in the massive weakly-annotated data is not fully exploited. 


\section{Method: Point Beyond Class}
To address the aforementioned challenges, we propose two regularization terms, namely multi-point consistency and symmetric consistency, and form a novel framework, \emph{i.e.,} Point Beyond Class (PBC), by integrating the two terms into Point DETR. The overview of our PBC is shown in Fig. \ref{fig:pipline}, where the proposed multi-point consistency and symmetric consistency are implemented to the step 1 and step 2 of original Point DETR, respectively.

\vspace{-0.2cm}
\subsection*{Step 1: Multi-Point Consistency for Box-Level Annotation}
As shown in Fig. \ref{fig:pipline}, the first step of our PBC is to train a teacher model with fully labeled data, where the model is trained to generate the bounding boxes from the point annotations. The Point DETR \cite{Chen_2021_CVPR} is adopted as backbone to process the input CXRs together with their point annotations. Denoted Point DETR as $\mathcal{F}_d(\cdot,\cdot)$, the process of bounding box prediction can be written as:
\begin{figure}[!t]
    \centering
    \includegraphics[width=.85\textwidth]{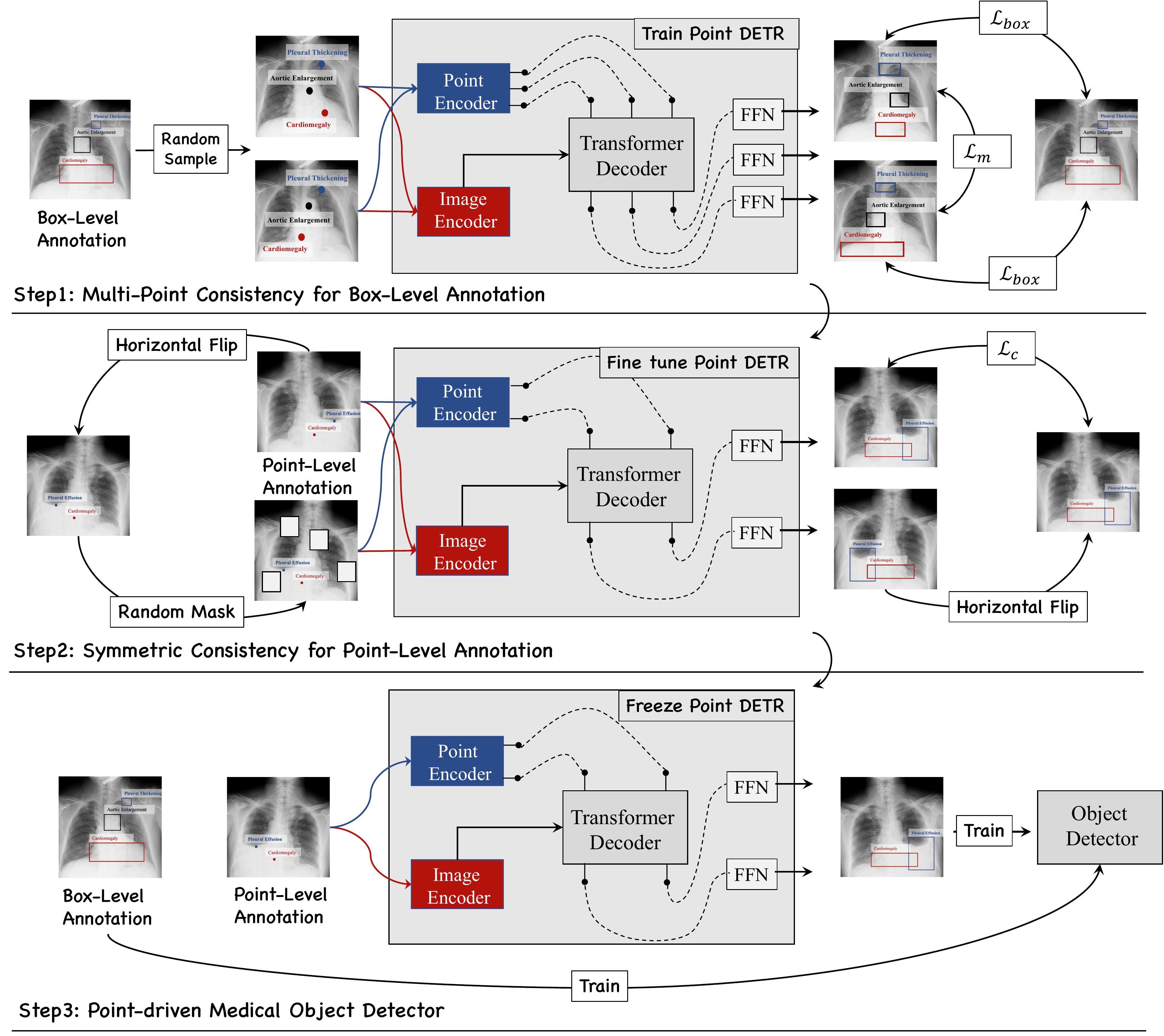}
    \caption{The pipeline of our proposed method, denoted as PBC, where annotations of different levels are processed separately by following a multi-step strategy.}
    \label{fig:pipline}
    \vspace{-3mm}
\end{figure}
\begin{equation}
    \mathcal{F}_d(\mathbb{X}_p,\mathbb{X}_i) = \hat{\mathbb{Y}},
\end{equation}
where $\mathbb{X}_p = (x, y, c)$ is the point annotation with $(x,y)$ for the positional information and $c$ for the abnormality category; $\mathbb{X}_i$ refers to the corresponding CXR and $\hat{\mathbb{Y}}$ is the predicted bounding box. The full objective $\mathcal{L}$ of this step is: 
\begin{equation}
    \mathcal{L} = \mathcal{L}_{box} + \mathcal{L}_m,
\end{equation}
where $\mathcal{L}_{box}$ and $\mathcal{L}_m$ are object detection loss and multi-point consistency loss, respectively. Here, we used the same object detection loss defined by DETR \cite{Chen_2021_CVPR,carion2020end} as $\mathcal{L}_{box}$. Due to the limited amount of fully-labeled data using in this step, the performance of Point DETR is sensitive to the position variation of point annotations (locating around center \emph{vs.} boundary of bounding box).
To mitigate this problem, we propose an auxiliary regularization term, \emph{i.e.,} multi-point consistency. As shown in  Fig. \ref{fig:pipline}, given $\mathbb{X}_i$ as an input image, two point annotations can be generated by randomly sampling inside the lesion-level bounding boxes (denoted as $\mathbb{X}^1_p$ and $\mathbb{X}^2_p$). Our multi-point consistency $\mathcal{L}_m$ aims to narrow down the distance between network predictions using $\mathbb{X}^1_p$ and $\mathbb{X}^2_p$, which can be formulated as:  
\begin{equation}
    \mathcal{L}_m = || \mathcal{F}_d(\mathbb{X}^1_p,\mathbb{X}_i) - \mathcal{F}_d(\mathbb{X}^2_p,\mathbb{X}_i) ||_2.
\end{equation}

\subsection*{Step 2: Symmetric Consistency for Point-Level Annotation}
In the previous methods, the point-level annotation is only employed to generate the pseudo lesion-level bounding box. The rich information contained in the weakly-annotated data is rarely exploited. In this regard, we propose a self-supervised constraint, termed symmetric consistency, to further refine the pseudo lesion-level labels, which enable the student model to learn the more accurate and robust feature representation.
Particularly, given a point-level annotation $\mathbb{X}_p$ and the corresponding image $\mathbb{X}_i$, we first perform the flipping operation $\mathcal{T}(\cdot)$ to them and obtain the flipped results ($\mathbb{X}^m_p$ and $\mathbb{X}^m_i$).
Then, we permute the content of original CXR by a random mask operator $\mathcal{M}(\cdot)$. The mechanism underlying our symmetric consistency is that the robust teacher model should be able to yield the consistent bounding box prediction under different image transformations. Hence,
taking ($\mathbb{X}_p$, $\mathcal{M}(\mathbb{X}^i)$) and ($\mathbb{X}^m_p$, $\mathbb{X}^m_i$) as input, the symmetric consistency can be defined as:
\begin{equation}
     \mathcal{L}_c = ||\mathcal{T}(\mathcal{F}_d(\mathbb{X}_p + \sigma ,\mathcal{M}(\mathbb{X}^m_i))) - \mathcal{F}_d(\mathbb{X}^m_p,\mathbb{X}^m_i) ||_2, 
\end{equation}
where $\sigma$ is a noise term sampling from a uniform distribution within $[-0.05,0.05]$ to prevent over fitting. 

\vspace{-0.2cm}
\subsection*{Step 3: Point-Driven Object Detector} 
After the above two steps, we get a well-trained Point DETR ($F_d(\cdot,\cdot)$), which is regarded as the teacher model to generate pseudo box labels for point-level weakly-annotated data. The generated pseudo labels can be employed to train the student model $\mathcal{F}_s(\cdot)$ using any modern detector, \emph{e.g.,} FCOS \cite{tian2019fcos} and Faster R-CNN \cite{ren2015faster,ren2016faster}, as backbone.
The training process can be written as:
\begin{equation}
    \min_{\mathcal{F}_s} \underbrace{\mathcal{L}_{o}(\mathcal{F}_s(\mathbb{X}_i),\mathcal{F}_d(\mathbb{X}_p,\mathbb{X}_i))}_{\text{Point-Level Annotation}} \quad +  \underbrace{\mathcal{L}_{o}(\mathcal{F}_s(\mathbb{X}_i),{\mathbb{Y}})}_{\text{Box-Level Annotation}}\!\!\!,
\end{equation}
where ${\mathbb{Y}}$ is the box-level annotation of $\mathbb{X}_i$, and $\mathcal{L}_{o}$ follows the loss function of $\mathcal{F}_s(\cdot)$ adopted by existing studies \cite{ren2015faster,ren2016faster,tian2019fcos}.

\vspace{-0.2cm}
\section{Experiments}
To validate the effectiveness of our PBC, extensive experiments are conducted on publicly-available datasets, including RSNA \cite{wang2017chestx} and VinDr-CXR \cite{nguyen2020vindr}. In this section, we first introduce the information of datasets and implementation details, and then construct the benchmarking for weakly semi-supervised abnormality localization in CXRs by comparing our PBC with the state-of-the-art weakly-supervised, semi-supervised and fully-supervised object detectors. Please note that, we have also tested our method on medical segmentation. More details can be found at our ArXiv Version.

\vspace{2mm}
\noindent{\bf Datasets.} The RSNA dataset came from a pneumonia detection challenge.\footnote{https://www.kaggle.com/c/rsna-pneumonia-detection-challenge/overview} The dataset consists of 26,684 CXRs, which can be categorized to negative or pneumonia. The lesion areas in pneumonia CXRs were identified and localized  by experienced radiologists. VinDr-CXR dataset,\footnote{https://vindr.ai/datasets/cxr} consists of 15,000 CXR images. The dataset providers invited experienced radiologist to annotate lesion areas of 14 thoracic diseases, \emph{e.g.,} aortic enlargement and cardiomegaly.

In this study, we separate each dataset to training and test sets according to the ratio of 80:20. Referring to the setting of weakly semi-supervised learning, we randomly sample 5\%, 10\%, 20\%, 30\%, 40\%, 50\% of training images as fully-labeled samples, while the rest only has the point-level annotation.\footnote{The strategy of point-level annotation generation is the same to \cite{Chen_2021_CVPR}.} Since the VinDr-CXR dataset has a long-tailed distribution, \emph{i.e.,} some abnormal categories only contain less than ten samples, we group eight categories with the least numbers of CXRs into one class (denoted as `Others') to stabilize the network training.

\vspace{2mm}
\noindent{\bf Implementation Details.} For a fair comparison, the network architecture of teacher model (\emph{i.e.,} our PBC) is consistent to \cite{Chen_2021_CVPR}.
The Adam optimizer is adopted for network optimization with an initial learning rate of $1\times10^{-4}$. 
Our PBC is observed to converge after 108 epochs of training. For the student detectors, we involve the widely-used FCOS and Faster R-CNN for evaluation. The student detectors are trained with stochastic gradient descent (SGD) optimizer. The model converges after 12 epochs of training. The setting of all hyper-parameters in the student model, including learning rate, weight decay and momentum, follows MMDetection \cite{chen2019mmdetection}.

\vspace{2mm}
\noindent{\bf Baselines and Evaluation Criterion.}
For the competing methods, we set the fully-supervised detector as the upper bound and point DETR \cite{Chen_2021_CVPR} without any regularization as the lower bound. In order to give a more comprehensive analysis of the proposed method, we also include a semi-supervised object detector \cite{sohn2020simple} for comparison. The mean average precision (mAP) is adopted as the evaluation metric. Note that we also evaluate several image-level-annotation-based weakly supervised approaches \cite{tang2018pcl} on the two datasets. However, due to the unclear boundaries of lesion areas, the region proposals are totally inaccurate, which results in an mAP $\leq$ 5\%. Hence, we do not include the results in the benchmark.

\subsection{Ablation Study}
To quantify the contribution of each regularization term in the proposed PBC, we evaluate the performance of the variants with/without each constraint. The evaluation results are presented in Table \ref{tab:ablation}. As shown, 
the proposed method outperforms the baseline (raw Point DETR) significantly with different numbers of box-level annotations. Using the multi-point consistency constraint, improvements of $+4.0\%$ and $+9.6\%$ can be achieved with 50\% fully labeled data on RSNA and VinDr-CXR datasets, respectively. The performance can be further boosted by using our symmetric consistency: the mAP reaches 39.2\% on RSNA and 28.5\% on VinDr-CXR, respectively, which surpasses the baseline by a large margin of $\sim$15\%.

\begin{table}[!t]
\caption{The ablation study carried on RSNA \cite{wang2017chestx}/VinDr-CXR \cite{nguyen2020vindr} based on Point DETR (teacher model) \cite{Chen_2021_CVPR} in the terms of mAP (\%).} 
\label{tab:ablation}
\resizebox{.98\textwidth}{!}{  
\begin{tabular}{ccc|c|c|c|c|c|c}
\hline
Baseline & Multi-Point & \begin{tabular}[c]{@{}c@{}}Symmetric \\ Consistency\end{tabular} & 5\%               & 10\%                & 20\%                & 30\%                & 40\%                & 50\%                \\ \hline
$\surd$       &$\times$                                                                &$\times$                                                             & 2.1/8.9          & 2.8/9.5            & 9.0/10.1           & 18.8/10.6          & 23.7/10.9          & 25.1/12.3          \\ \hline
$\surd$       & $\surd$                                                                &$\times$                                                             & 3.4/9.4          & 9.5/9.3            & 18.3/13.4          & 23.8/15.8          & 25.6/20.1          & 29.1/21.9          \\
$\surd$       &$\times$                                                                & $\surd$                                                             & 4.1/8.0          & 8.4/9.6            & 14.8/9.9           & 24.6/10.5          & 30.6/12.4          & 32.4/11.8          \\ \hline 
\rowcolor[HTML]{EFEFEF}
$\surd$       & $\surd$                                                                & $\surd$                                                             & \textbf{6.8/9.5} & \textbf{17.8/13.0} & \textbf{28.1/15.4} & \textbf{34.4/21.9} & \textbf{36.9/27.5} & \textbf{39.2/28.5} \\ \hline
\end{tabular}
}
\end{table}
\begin{table}[!t]
\caption{mAP (\%) on RSNA \cite{wang2017chestx}/VinDr-CXR \cite{nguyen2020vindr} based on different student detectors trained with various methods.}
\label{tab:comparison}
\resizebox{.98\textwidth}{!}{  
\Large
\begin{threeparttable}[b]
\begin{tabular}{c|c|c|c|c|c|c|c|c}
\hline
Detector                                                                  & Method                             & 5\%                                         & 10\%                                        & 20\%                                        & 30\%                                        & 40\%                                        & 50\%                                        & 100\%                        \\ \hline
                                                                          & Only Box                           & 14.1/0.9                                   & 22.5/19.2                                  & 29.7/15.9                                  & 33.1/19.7                                  & 33.4/24.1                                  & 34.4/29.9                                  &                             \\ \cline{2-8}
                                                                           & Weak-Sup.$^*$         &                 \multicolumn{6}{c|}{$\leq$5\%} &                             \\\cline{2-8}
                                                                          & Semi-Sup.                          & 17.4/1.4                                   & 24.2/14.9                                  & 32.0/25.0                                  & 36.5/24.1                                  & 37.8/31.7                                  & 38.9/33.5                                  &                             \\
                                                                          & Point DETR                         & 21.8/19.4                                  & 28.3/24.9                                  & 34.4/27.1                                  & 37.3/27.7                                  & 38.5/32.3                                  & 39.8/34.7                                  &                             \\ \cline{2-8}
\multirow{-4}{*}{FCOS \cite{tian2019fcos}}                                                    & \cellcolor[HTML]{EFEFEF}PBC (Ours) & \cellcolor[HTML]{EFEFEF}\textbf{25.8/21.8} & \cellcolor[HTML]{EFEFEF}\textbf{32.4/25.1} & \cellcolor[HTML]{EFEFEF}\textbf{37.1/29.9} & \cellcolor[HTML]{EFEFEF}\textbf{40.5/31.3} & \cellcolor[HTML]{EFEFEF}\textbf{40.2/34.4} & \cellcolor[HTML]{EFEFEF}\textbf{42.6/36.4} & \multirow{-4}{*}{43.1/39.2} \\ \hline\hline
                                                                          & Only Box                           & 14.0/2.0                                   & 17.0/13.3                                  & 24.9/27.4                                  & 29.9/31.2                                  & 33.4/34.4                                  & 35.6/35.7                                  &                             \\ \cline{2-8}
                                                                                                                                                     & Weak-Sup.         &                 \multicolumn{6}{c|}{$\leq$5\%} &                             \\\cline{2-8}
                                                                          & Semi-Sup.                          & 14.6/13.5                                  & 19.5/21.5                                  & 30.4/29.9                                  & 33.7/32.0                                  & 37.3/34.6                                  & 39.5/36.5                                  &                             \\
                                                                          & Point DETR                         & 17.4/21.8                                  & 19.4/25.2                                  & 31.7/29.8                                  & 38.4/31.9                                  & 39.2/34.7                                  & 40.8/36.2                                  &                             \\ \cline{2-8}
\multirow{-4}{*}{\begin{tabular}[c]{@{}c@{}}Faster \\ R-CNN \end{tabular}} & \cellcolor[HTML]{EFEFEF}PBC (Ours) & \cellcolor[HTML]{EFEFEF}\textbf{23.3/23.3} & \cellcolor[HTML]{EFEFEF}\textbf{30.4/26.7} & \cellcolor[HTML]{EFEFEF}\textbf{37.7/31.9} & \cellcolor[HTML]{EFEFEF}\textbf{39.3/33.7} & \cellcolor[HTML]{EFEFEF}\textbf{40.7/36.2} & \cellcolor[HTML]{EFEFEF}\textbf{43.1/37.2} & \multirow{-4}{*}{43.0/39.5} \\ \hline
\end{tabular}
\begin{tablenotes}
\item $^*$ Since there is less contextual information contained in CXRs, the re-implemented WSOD method (PCL\cite{tang2018pcl}) cannot obtain competitive results. More details can be found in appendix. 
\end{tablenotes}
\end{threeparttable}
}
\vspace{-0.3cm}
\end{table}

\subsection{\bf Performance Benchmark}
To construct the benchmark for weakly semi-supervised abnormality localization with CXRs,
the performance of our PBC method is compared with approaches under different training strategies on RSNA and VinDr-CXR datasets. As shown in Table \ref{tab:comparison}, our PBC method consistently outperforms other methods on both datasets. Concretely, with 30\% full labeled data, the FCOS trained with the proposed PBC method achieves an mAP of $40.5\%$, which is comparable to the one with 100\% fully-labeled data ($43.1\%$).
The experimental results indicate that the proposed PBC method can balance the trade-off between annotation cost and model accuracy. Furthermore, compared to the listed semi-supervised and weakly semi-supervised methods, our PBC yields more significant improvements to box-only models, especially with extremely limited fully-labeled data. Specifically, the proposed method boosts the mAP of Faster R-CNN to $23.3\%$ on both datasets with only 5\% fully-labeled samples, \emph{i.e.,} $+5.9\%$ and $+1.5\%$ higher than the runner-up (Point DETR), which demonstrates the effectiveness of our regularization terms on refining the pseudo bounding boxes for the student model.

\section{Conclusion}
In this paper, we constructed a benchmark for weakly semi-supervised abnormality localization with CXRs. A novel framework, namely point beyond class (PBC), was formed, which consists of two novel regularization terms (multi-point consistency and symmetric consistency). In particular, our multi-point consistency drives the model to localize the consistent bounding boxes from different points inside the same lesion area. While, the proposed symmetric consistency enforces the network to yield consistent predictions for the same CXR permuted by different transformations. These two regularization terms thereby improve the robustness of learned features against the variety of point annotations. The effectiveness of the proposed PBC method has been validated on two publicly available datasets, \emph{i.e.,} RSNA and VinDr-CXR.

\subsubsection{Acknowledgment} This work was supported in part by the National Natural Science Foundation of China (Grant No. 91959108), Key-Area Research and Development Program of Guangdong Province, China (No. 2018B010111001), National Key R\&D Program of China (2018YFC2000702) and the Scientific and Technical Innovation 2030-"New Generation Artificial Intelligence"  Project (No. 2020AAA0104100).

\bibliographystyle{splncs04}
\bibliography{cite}
\end{document}